# Labeling Case Similarity based on Co-Citation of Legal Articles in Judgment Documents with Empirical Dispute-Based Evaluation

Chao-Lin Liu     Po-Hsien Wu     Yi-Ting Yu

National Chengchi University, Wen-Shan, Taipei 116011, Taiwan
`{chaolin, 111753120, 112752502}@nccu.edu.tw`

**Abstract.** This report addresses the challenge of limited labeled datasets for developing legal recommender systems, particularly in specialized domains like labor disputes. We propose a new approach leveraging the co-citation of legal articles within cases to establish similarity and enable algorithmic annotation. This method draws a parallel to the concept of case co-citation, utilizing cited articles as indicators of shared legal issues. To evaluate the labeled results, we employ a system that recommends similar cases based on plaintiffs' accusations, defendants' rebuttals, and points of disputes. The evaluation demonstrates that the recommender, with finetuned text embedding models and a reasonable BiLSTM module can recommend labor cases whose similarity was measured by the co-citation of the legal articles. This research contributes to the development of automated annotation techniques for legal documents, particularly in areas with limited access to comprehensive legal databases.

## 1 Introduction

Legal recommender systems are frequently developed to assist lawyers and judges in identifying relevant or similar cases, e.g., [16], offering valuable insights for drafting decisions and predicting case outcomes. Analysis of past cases can also contribute to recidivism prediction, informing sentencing and rehabilitation strategies [12]. Furthermore, in labor disputes, access to similar cases can benefit both laborers and employers, and may improve the efficacy of mediation, particularly in jurisdictions where mediation is a prerequisite to litigation. By facilitating pre-court resolution, such systems can potentially alleviate the burden on the judicial system.

The development of machine learning-based recommender systems often necessitates substantial training data. While the availability of such data can vary depending on the research objectives, for example, if the goal is to categorize cases by the specific charges involved, there might be hundreds of thousands of examples for certain types of charges [9]. Many studies have relied on significantly smaller datasets, with some utilizing only a few thousand or even as few as one thousand cases [10].

Several substantial open databases exist for legal research. Notably, Wang et al.

utilized the Supreme Court Database (SCDB), which encompasses case metadata from 1946 to 2020, comprising information on 7,800 cases [18]. Furthermore, CAIL offers a collection of 8,138 annotated Chinese cases [1], while MUSER provides another public, annotated Chinese dataset of over 4,000 cases specifically designed for similar case retrieval [7].

The scarcity of resources for legal recommender systems is partly attributable to the broad and often ambiguous definitions of "similarity" and "relevance" in legal contexts. Determining what constitutes "similarity" can be a complex legal issue [17], as the interpretation of legal language is highly domain-specific and language-dependent. This complexity necessitates expert annotation, which is expensive and difficult to obtain at scale. Furthermore, even when such annotations exist, they are rarely made publicly available [11]. The concept of "similarity" can also be highly specific, as cases are only considered relevant if they directly support the judgment of the current case, often requiring analysis at the paragraph level [14]. These challenges explain why empirical studies in this domain frequently utilize only a few thousand cases.

Algorithmic labeling of legal data presents a promising solution to the challenge of limited labeled datasets. Researchers are exploring machine learning and active learning techniques to automate this process and generate larger, more readily available datasets for legal research and analysis [15].

Leveraging existing information within legal documents for automated labeling presents a promising avenue for generating large-scale annotated legal datasets. This approach, exemplified by studies on co-citation networks and bibliographic coupling, posits that documents citing the same references are inherently related, enabling algorithmic identification of relevant documents [3, 11]. By programmatically labeling cases based on these inherent relationships, substantial datasets can be generated, with human experts subsequently sampling and inspecting the labeled data to ensure quality assurance.

This study examines the use of co-citation of legal articles as a basis for establishing similarity between labor dispute cases, drawing a parallel to the concept of co-citation of previous cases. While labor dispute cases are already a specialized area, nuanced differences exist within subcategories. Co-citation of legal articles in past cases provides valuable insights into the relationships between them. By citing articles, judges implicitly acknowledge relevant disputes, effectively using these citations to encapsulate the core issues within labor dispute cases.

To evaluate the efficacy of this proposed method, we conduct a practical experiment. Specifically, we employ a system that recommends similar cases based solely on the plaintiff's accusations, the defendant's rebuttals, and the identified points of disputes. If these recommendations accurately predict similar cases as determined by the co-citation analysis of legal articles, it will strongly support the viability of this approach for algorithmic annotation.[1] This finding would have significant implications for automating the annotation of legal documents, particularly in specialized domains like labor disputes and in low-resource settings where access to comprehensive legal databases may be limited.

---

[1] As researchers in legal informatics, we cannot underestimate the importance and diversity of human preferences and judgments. Here, "algorithmic annotation" refers to an alternative of expert annotation, particularly when human annotation for a large dataset is difficult to achieve. This is a response to a reviewers' comment.

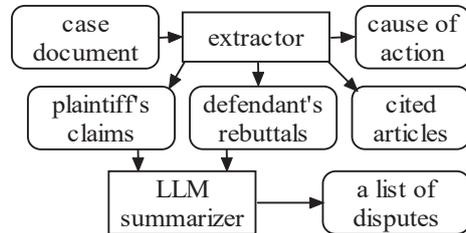

Fig. 1. Extracting and preprocessing information from the judgment documents

Our evaluation demonstrated the effectiveness of this approach. By finetuning a text embedding model on labeled cases, we achieved improved performance in a similar case recommendation task compared to using only a pretrained model. Furthermore, a BiLSTM-based model yielded promising results in recommending similar labor dispute cases. The quality of these recommendations suggests the feasibility of expanding the system to provide justifications and explanations, thereby enhancing its practical value for legal professionals.

This report begins with a brief literature review in the Introduction. Section 2 introduces the core concept of labeling case similarity via legal article co-citation, while Section 3 explores alternative similarity metrics. Sections 4 and 5 detail the methodology and data sources used to evaluate the reliability of this co-citation-based similarity. Sections 6 and 7 present a two-stage evaluation: first, by fine-tuning a text embedding model using the labeled data, and second, by assessing a BiLSTM-based model's ability to recommend similar labor dispute cases based on litigant arguments. Finally, Section 8 addresses further technical considerations and concludes the report.

## 2 Co-Citation-based Labeling

We first utilize a dataset of 2,886 judgment documents of labor disputes that were sourced from the Taiwan Judicial Yuan Open Database (**TJYOD**) to illustrate the main idea of labeling case similarity with co-citation of legal articles. These cases were adjudicated in Taiwanese district courts between 2020 and 2023, with the following distribution: 744 cases in 2020, 707 in 2021, 809 in 2022, and 626 in 2023 [20].

Each judgment document contains comprehensive information, including sections for the *plaintiffs' claims*, the *defendants' rebuttals*, the *cited legal articles*, and the *cause of action* (sometimes called *title*), as illustrated in Fig. 1. Although the organizations of the judgment documents were not prepared for machine learning, the formats of most of the real judgment documents are reasonable and usable, and we can apply natural language processing (NLP) methods, including regular expressions and syntactic analysis, to extract the statements of the litigants' claims and the cause of action. In addition, we have to rely on the techniques of *selenium*[2] to extract the cited legal articles from the web page of each individual judgment documents.

---

[2] Many online tutorials for selenium are available, e.g., https://www.scraperapi.com/web-scraping/selenium/.

While all four data elements are utilized in this research, this section focuses specifically on the cited legal articles. We present a preliminary analysis of these cases and detail our methodology for using them in the labeling and evaluation processes.

### 2.1 Legal Articles and Codes Cited in the Case Collection

Analysis of the judgment documents obtained from TJYOD reveals that most of them contain information regarding cited legal articles. A preliminary examination of these documents indicates that 1381 distinct legal articles from 153 different codes (including acts, regulations, rules, etc.) were cited across the 2886 cases.

Fig. 2 illustrates the distribution of the number of cited legal articles. The vertical axis represents the number of cases, while the horizontal axis represents the number of distinct legal articles cited per case. The category "25+" denotes cases citing at least 25 distinct legal articles. On average, each case cited 10.38 distinct legal articles. Notably, 29 cases cited more than 30 distinct articles. Excluding these more complex cases, the average number of distinct legal articles cited per case is 10.18.

Another analysis focuses on the distinct legal codes cited in each case (e.g., civil, criminal, or labor), rather than the total number of articles cited within each code. For instance, a case citing multiple articles from only the civil code would be categorized as "1," while a case citing articles from both the civil and labor codes would be categorized as "2."

Fig. 3 illustrates the distribution of cited codes within the dataset. The average number of codes cited per case is 3.61, with a maximum of 12 codes in the most complex case. Excluding cases citing eight or more codes (only 60 cases in total), the average drops slightly to 3.51. This indicates that the majority of cases cite articles from a relatively limited number of legal codes.

As expected, the most frequently cited codes in the labor dispute cases is the *Labor Standards Law* (勞動基準法, hereafter "LSL"), followed by the *Civil Law* (民法), the *Labor Pension Act* (勞工退休金條例), the *Enforcement Rules of the Labor Standards Act* (勞動基準法施行細則, hereafter

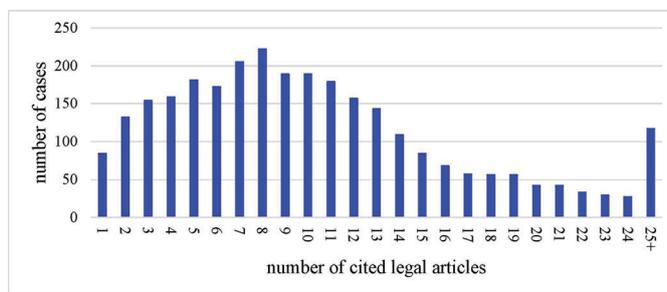

**Fig. 2.** distribution of the number of cited legal articles

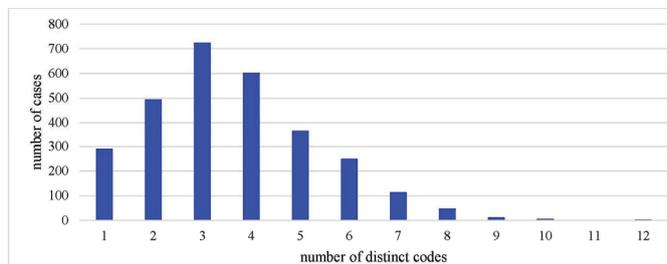

**Fig. 3.** distribution of the number of cited codes

"**ERLSA**"), and the *Labor Occupational Accident Insurance and Protection Act* (勞工職業災害保險及保護法).

Our analysis revealed that the cases cited articles from 153 different laws and regulations in Taiwan. Many of these pertain to specific industries, such as the *Seafarer Law* (船員法), the *Law of Pharmacists* (藥師法), and the *Teachers' Act* (教師法, hereafter "**TA**").

The most cited legal provisions are Article 2 of the LSL, followed by Article 11 of the LSL, Article 55 of the LSL, Article 229 of the Civil Law, Article 12 of the LSL, Article 10 of the ERLSA, and Article 12 of the Labor Pension Act.

If we have the luxury to investigate the details of individual cases by reading the contents of the judgment documents, we could understand why some specific provisions were frequently cited. For instance, the Article 2 of the LSL is about the definition of "laborers", and a lawsuit may begin with the establishment of whether the plaintiffs were laborers in terms of law. The frequent applications of Article 55 of the LSL were related to the changes in law about the calculation of pension in recent years. These are reasonable supporting factors for considering whether cases are similar.

Only a small portion of judgment documents directly contained a list of disputes between the litigants, due to the legal processing for handling the labor disputes. Although it is possible to apply more complex NLP methods to identify the disputes from the judgment documents, we relied on large language models (LLMs) to produce the disputes. We used GPT-4-Turbo-0409 in the work reported in this paper. More details for obtaining the disputes with LLMs will follow in Section 4.1.

### 2.2 Labeling Similarity Based on Co-Citations

The cited legal articles within the case documents are official records, thus providing arguably reliable annotations for each case by legal experts. Furthermore, it is logical to infer that judges base their rulings on the facts of the case in conjunction with relevant legal articles, with the exception of procedural codes, such as the *Code of Civil Procedure* (民事訴訟法).

Therefore, utilizing cited articles as a reflection of the disputes and facts within the cases is a reasonable approach. While alternative methods for measuring case similarity must exist [17], measuring similarity based on co-citations of legal articles presents a valid choice. Based on this premise, we employ the co-citations of legal articles to determine the similarity between two cases. This definition of similarity serves as the foundation for labeling cases as similar or dissimilar.

To assess the similarity between cases based on their cited legal articles, we utilize the DICE coefficient [2]. This metric quantifies the similarity between two sets, in this context, the sets of legal articles cited by each case. Let $A_i$ represent the set of legal articles cited by case $C_i$. The DICE coefficient between the sets of legal articles cited by cases $C_i$ and $C_j$ is calculated as follows:

$$\text{DICE}(C_i, C_j) = \frac{2|A_i \cap A_j|}{|A_i|+|A_j|} \quad (1)$$

where $|A_i|$ denotes the number of distinct elements (legal articles) in set $A_i$ and $|A_i \cap A_j|$ the number of distinct elements (legal articles) common to both $A_i$ and $A_j$.

This coefficient provides a normalized measure of similarity, ranging from 0 (no shared articles) to 1 (identical sets of cited articles).

A significant limitation of the original DICE definition in (1) is its tendency to overestimate similarity between cases with fewer co-cited legal articles. To address this, we can introduce a weighting mechanism to the numerator in (1).

A straightforward heuristic approach is to establish a minimum threshold, $\alpha$, for the desired number of co-cited articles. This threshold can then be incorporated into a modified DICE calculation, as shown in (2). For instance, if $\alpha = 3$, DICE2 would become a lower similarity score than DICE for case pairs with fewer than three co-cited articles, and a higher score for pairs exceeding this threshold.

$$\text{DICE2}(C_i, C_j) = \frac{2|A_i \cap A_j|}{|A_i|+|A_j|} \times \frac{|A_i \cap A_j|}{\alpha} \tag{2}$$

The selection of $\alpha$ may depend on the needs of the applications, the laws in a nation, and the human feedback in user studies.

## 3 Encoding the Citations of Legal articles

The method used to encode cases based on cited legal articles should accurately reflect our preferred criteria for determining similarity. To illustrate this complex issue, let's consider a simplified example with a query case, $Q$, and a candidate case, $X$.

Assume that case $Q$ cites articles lsl_1 and lsl_3 from the LSL (*Labor Standards Law*) and articles ta_5 and ta_7 from TA (*Teachers' Act*). Furthermore, assume that case $X$ cites articles lsl_1 and lsl_2 from the LSL and articles ta_5 and ta_6 from TA.

### 3.1 Simple and Direct Encoding

A common approach to measuring the similarity between case $Q$ and case $X$ is to calculate the DICE coefficient between their sets of cited legal articles [4]. Let $A_Q$ and $A_X$ represent the sets of legal articles cited in case $Q$ and case $X$, respectively. The Dice coefficient between $A_Q$ and $A_X$ is defined in Equation (1).

For example, we can encode $A_Q$ as {lsl_1, lsl_3, ta_5, ta_7} and $A_X$ as {lsl_1, lsl_2, ta_5, ta_6}. Applying Equation (1) to cases $Q$ and $X$ yields DICE($Q$, $X$)=(2×2)/(4+4)=.5.

### 3.2 Perspective Encoding

While encoding cases with all cited legal articles is conceivable practice, specific applications may prefer tailored approaches to defining "similarity." For instance, a legal expert might prioritize similarities in citations from the *Labor Standards Law* (LSL) over those from the *Teachers' Act* (TA), without entirely discounting shared TA citations.

In such a scenario, the expert might encode cases $Q$ and $X$ as {lsl_1, lsl_3, TA} and {lsl_1, lsl_2, TA}, respectively. This encoding retains all LSL articles while using "TA" as a general indicator of one or more citations from the Teachers' Act. This approach would yield a Dice coefficient of DICE($Q$, $X$) = 2×2/(3+3) = 2/3. It is important to note that this method increases the perceived similarity by overlooking the specific differences in TA articles cited by Q and X.

Therefore, "TA" functions as a **grouping feature** in this example. Grouping features allow encoders to disregard specific distinctions between lower-level features during encoding. It is important to note that a grouping feature can also represent an individual legal article, such as lsl_1 or lsl_3.

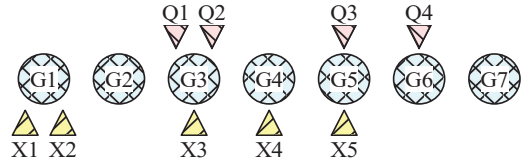

**Fig. 4.** A simple illustration of generalized DICE

A sophisticated encoding may even consider and explore a hierarchical structure of laws to optimally capture the subtle differences between different levels of similarity, where "different levels of similarity" may depend on the cultural and legal concepts and traditions of different nations. For labor and employment relationships in Taiwan, severance payment, retirement pension, salary, improper layoff, and overtime pay are relatively important and common issues. Hence, one my use them as grouping features in a hierarchical structure, where codes and their chapters form two or more levels.

### 3.3 A Generalized DICE Coefficient

This subsection explores refinements to the DICE coefficient. Consider the illustrative example in Fig. 4, where circles represent user-defined grouping features. For simplicity, Fig. 4 depicts only seven grouping features (G1 through G7), though in practice, given our dataset of 1381 articles from 153 codes (see Section 2.1), hundreds or even thousands of grouping features may exist.

Each small triangle in the figure symbolizes an individual legal article cited within a case. Inverted triangles ($\triangledown$) represent articles cited in case $Q$ (Q1, Q2, Q3, and Q4), while upward triangles ($\triangle$) represent articles cited in case $X$ (X1, X2, ⋯, X5). Colors and varied slash patterns are employed in Fig. 4 to enhance visual clarity and differentiation.

In this example, cited articles in case $Q$ appear within three grouping features (G3, G5, and G6), whereas those cited in case $X$ are distributed across four (G1, G3, G4, and G5). The triangles representing cited articles point to their corresponding grouping features; for instance, Q1, Q2, and X3 all belong to G3. Notably, $Q$ and $X$ share citations within two groups: G3 and G5. Using the method of Liu and Liu [8], the similarity score between $Q$ and $X$ would be 2.

To refine this similarity score, we reconsider how legal articles are co-cited within $Q$ and $X$. Let $S \sqcap T$ denote the set of articles co-cited by cases $S$ and $T$ and combinations of the co-citations. More specifically, $Q \sqcap X = \{(Q1, X3), (Q2, X3), (Q3, X5)\}$.

We propose a generalized DICE coefficient in Equation (3) to measure the co-citation similarity between cases $Q$ and $X$. Here, $s(A_i \sqcap A_j)$ denotes the total number of distinct articles participating in the intersection $A_i \sqcap A_j$, calculated by summing the number of distinct articles contributing to the intersection from each set.

$$g\_DICE(A_i, A_j) = \frac{s(A_i \sqcap A_j)}{|A_i|+|A_j|} \quad (3)$$

In the example illustrated in Fig. 4, three articles (Q1, Q2, and Q3) cited by $Q$ participate in the intersection $Q \sqcap X$, and two articles (X3 and X5) cited by $X$ participated in the intersection. Therefore, $s(Q \sqcap X) = 3 + 2 = 5$ and $g\_DICE(Q,X) = 5/(4+5) = 5/9$. For the example that we discussed in Section 3.2, TA is a grouping feature, so $Q \sqcap X = \{(lsl\_1, lsl\_1), \{(ta\_5, ta\_5), (ta\_5, ta\_6), (ta\_7, ta\_6)\}\}$ and $g\_DICE(Q,X) = (3+3)/(4+4) = 3/4$.

### 3.4 Distributions of the Similarity Scores

In Equation (2), we assigned a value of 3 to the α parameter to prioritize co-citations. Furthermore, to establish the grouping features for the $g\_DICE$ score calculation in Equation (3), we first limited our analysis to statutes pertaining to labor laws, such as the *Labor Standards Law*.

Our analysis encompassed over 4 million case pairs derived from the 2,886 cases in our dataset. As anticipated, the $g\_DICE$ scores for most case pairs were negligible, with over 70% falling below 0.1. Consequently, this range is not shown in Fig. 5.

Fig. 5 illustrates the distribution of $g\_DICE$ scores greater than or equal to 0.1. The category "0.3+" represents scores between 0.3 and 0.4, while "1.5+" denotes scores greater than or equal to 1.5. Case pairs exhibiting such high similarity (1.5+) constituted less than 0.1% of the total. In fact, only approximately 1.27% of case pairs had $g\_DICE$ scores exceeding 1.0 when α = 3.

Fig. 6 shows the distribution of the $g\_DICE$ coefficients when we considered the citations of all of the cited legal articles and without introducing any grouping features. Legal articles of special codes, e.g., the *Teacher's Act* and *Law of Pharmacists* were included. The percentage of case pairs whose similarity is larger than 1.0 increase to more than 2.70%.

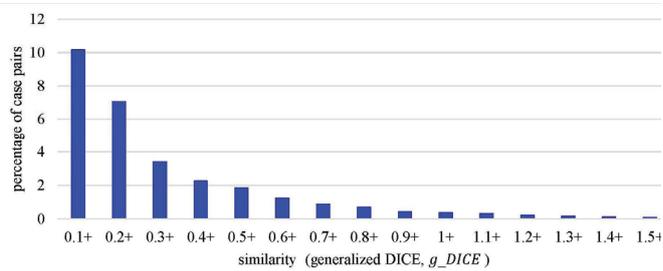

**Fig. 5.** distribution of the similarity scores when considering only

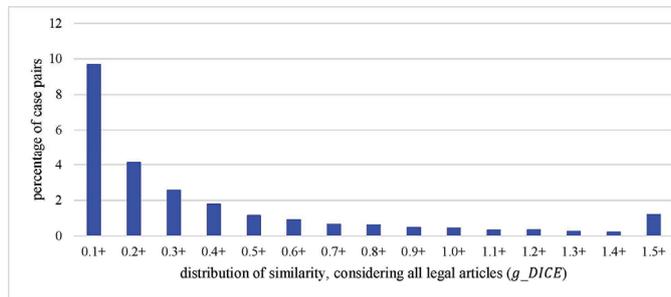

**Fig. 6.** distribution of the similarity scores when considering all cited legal articles ($g\_DICE$)

The choice of the different DICE-based coefficients and the inclusions of grouping features clearly influence the distributions of the coefficients and the resulting categorization of similarity. We focus on the labor related laws and set a to 3, so we used a $g\_DICE$ scores in the evaluation section (Sections 6 and 7).

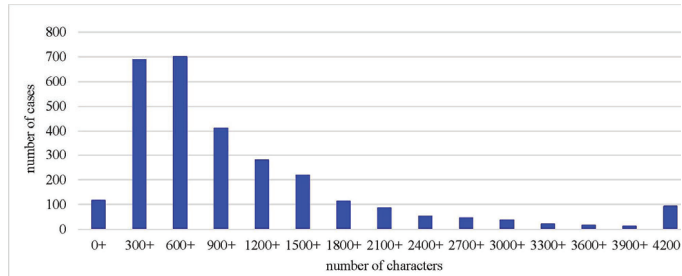

**Fig. 7.** distribution of the length of plaintiff's accusations

Regardless of the specific DICE coefficient definition employed in an analysis, we recommend prioritizing candidate cases with higher DICE coefficients in relation to the query case.

## 4 Reliability of the Co-Citation-based Labels

To assess the usability of the co-citation-based labels for similar case recommendations (SCR), we can construct an SCR recommender system that relies exclusively on the disputed issues between plaintiffs and defendants. If the recommendations generated by this system exhibit a substantial degree of agreement with the co-citation-based labels, this would support the practical utility and reliability of the labeling method.

### 4.1 Itemizing the Disputes

We extracted the plaintiff's accusations and the defendant's rebuttals from a judgment document to calculate the number of characters in the accusations and the rebuttals. Fig. 7 illustrates the distribution of accusation lengths. The horizontal axis represents character count ranges (e.g., "300+" denotes the range 300 to 599 characters). The vertical axis represents the corresponding number of cases. Notably, almost 100 cases involved accusations exceeding 4,200 characters, with the shortest being 73 characters and the longest exceeding 10,000. The average accusation length was 1,235 characters. Excluding the longest 10% of accusations reduces this average to 923 characters.

While modern text embedding models ca n handle text exceeding 512 characters, concerns remain about whether one embedding can fully capture the nuances of lengthy legal arguments. To address this concern, we employed large language models (LLMs), specifically GPT-4-Turbo-0409 (hereafter "GPT"), to summarize and itemize accusations and rebuttals, subsequently asking the GPT to identify and itemize the points of dispute. Summarization is one of the LLMs services that gain higher confidence among the public, the degree the confidence extends to the realm of law is improving but remain undecidable, particularly for Chinese legal texts [13].

To mitigate potential inaccuracies ("hallucinations"), we adopted a chain-of-thought (COT) approach [19]. This involved sequentially prompting GPT to: (1) summarize and itemize accusations; (2) summarize and itemize rebuttals; and (3) identify and itemize disputes absolutely based on the itemized accusations and rebuttals. A low-temperature setting was used throughout to ensure focused and consistent responses.

For the 2,886 judgment documents, the number of dispute items per case ranged from 1 to 10, with an average of 3.93. The average total character count in the dispute items was 102.75, with a maximum of 225 characters in the most complex case.

### 4.2 Embedding the Itemized Disputes

We used the itemized disputes to represent a given case, and compared the itemized disputes of two cases to determine their similarity when recommending similar cases.

To achieve this, we employed publicly available embedding models, specifically `multilingual-e5-large`[3] (Hugging Face) and `text-embeder-3-large`[4] (OpenAI), to transform textual dispute statements into numerical vector representations. These embeddings facilitated the development of both clustering-based and classification-based methodologies for identifying similar cases.

## 5 Three Disjoint Casesets for Empirical Evaluation

To demonstrate the utility of our aforementioned labeling procedures, we applied them to a larger collection of labor dispute cases and subsequently used this labeled data to finetune a text embedder for similar case recommendations.

In order to mitigate the risk of data leakage stemming from related cases, our dataset was restricted to judgments from district courts, e.g., excluding any judgments for appeal cases from higher courts. Furthermore, we partitioned the data into three mutually exclusive sets: one for finetuning the text embedder, one for training the models for similar case retrieval (SCR) tasks, and one for evaluating the SCR results. This partitioning was crucial to prevent the finetuned embedder from having prior exposure to the cases used in the SCR tasks, ensuring the integrity of our evaluation.

For finetuning the text embedder, we utilized 3,380 judgment documents to enhance the performance of the `multilingual-e5-large` text embedder. This model was selected due to its superior performance with traditional Chinese texts, as demonstrated in published evaluations and our internal assessments.[5]

Another set of 2,630 cases were allocated for training and validating the models for the SCR tasks. Following the finetuning of the embedder, we generated dispute lists for these cases using the procedures outlined in Section 4. In essence, we represented each labor dispute case in the SCR process by the itemized disputes between the plaintiffs and defendants, without incorporating any information about cited legal articles. Our proposed annotation and evaluation procedures are validated if the recommender can identify cases that cited a similar set of legal articles as the query case.

Finally, we reserved a smaller collection of 499 cases for the final evaluation. We employed the same similarity calculation method across all 124,251 pairs within this collection.[6] A subset of these 499 cases will serve as query cases to evaluate the recommender's ability to identify similar cases from the remaining pool.

---

[3] Hugging Face: https://huggingface.co/intfloat/multilingual-e5-large
[4] OpenAI: https://platform.openai.com/docs/guides/embeddings
[5] https://ihower.tw/blog/archives/12167 (in Chinese)

## 6 Evaluation Step 1: Finetuning Text Embedders

We finetuned text embedders using labeled cases and a supervised contrastive learning method [6]. Subsequently, we employed both pretrained and finetuned embedders in a simplified version of the SCR task to assess whether finetuning enhanced the embedder's performance.

### 6.1 Preparing the Training Data

We may build our training dataset with the following steps.
1. Calculate the similarity between the cases in our collection of cases as we discussed in Section 3, and create "similar" and "not-similar" case pairs.
2. Sample the annotated case pairs with a plan (explained below).
3. Embed the dispute items of the sampled case pairs with the text embedder model `text-embedding-3-large` of OpenAI.
4. Select $m$ pairs of dispute pairs from the sampled case pairs, and add them into the training set for finetuning the embedder.

We may then use this trainset to finetune the text embedder.

At step 1, we continued to use the steps for computing the similarity scores that we used to generate the data for Fig. 5 in Section 3.4. We have a total of 5,710,510 case pairs, so we have a huge pool to choose.[7] Based on our discussion in Section 3.4, we treated cases pairs whose $g\_DICE$ are larger than 1.0 as "similar cases." At the stage of training the text embedder, we aimed to offer data of higher quality. Therefore, we did not use any case pairs whose similarity scores are between 0.5 and 1.0. Case pairs whose similarity scores are smaller than 0.5 were treated as "not-similar."

Using case pairs with relatively extreme similarity values, e.g., below 0.5 and above 1.0, helped the text embedder to learn to distinguish statement pairs from similar and not-similar cases and to improve its embedding quality. Differences between the disputes in cases in a pair with mid-range similarity, [0.5, 1], can be less clear even for human annotators and can be confusing for algorithmic learners, i.e., the text embedder in this context.

At step 2, we randomly sampled "similar" and "not-similar" case pairs from the pool. We created 50,000 sample instances for training the text embedder. To better train the embedder, we intentionally sampled 60% from the not-similar case pairs and 40% from the similar case pairs. If we sampled the case pairs randomly, we would find much fewer statement pairs from similar case pairs, as we showed in Figures 5 and 6.

It seems puzzling that we relied on the help of `text-embedding-3-large` to train another embedder. Assume that cases α and β are a sampled pair, and assume that α and β have $p$ and $q$ dispute items respectively. Should we use all $pq$ combinations of dispute pairs to train the embedder?

---

[6] Choosing any two cases from 499 cases have 499×498÷2 = 124,251 combinations.
[7] Choosing any two cases from 3380 cases have 3380×3879÷2 = 5,710,510 combinations.

It may be logically problematic to assume that all dispute pairs from a pair of cases that are considered similar based on their co-citation of legal articles. Hence, if we want to use only $m$ (smaller than $pq$) of these $pq$ combinations in the training data at step 4, we must have a way to choose. This is when we needed the OpenAI embedder.

At step 4, we calculate the Euclidean distance of the embeddings (returned from `text-embedding-3-large`) for all of the $pq$ combinations. If the current pair is "similar", we choose the $m$ pairs that have the smallest Euclidean distance. If the current pair is "not-similar", we choose the $m$ pairs that have the largest Euclidean distance.

**Table 1.** Finetuning improved the text embedder

| FT? | topN | precision | NDCG |
|---|---|---|---|
| no | 5 | 71.21% | 0.6969 |
| yes | 5 | 74.29% | 0.7486 |
| no | 10 | 68.02% | 0.6697 |
| yes | 10 | 72.31% | 0.7232 |
| no | 15 | 67.99% | 0.6677 |
| yes | 15 | 69.38% | 0.6959 |
| no | 20 | 68.13% | 0.6642 |
| yes | 20 | 70.22% | 0.6845 |
| no | 25 | 66.81% | 0.6553 |
| yes | 25 | 70.90% | 0.6933 |
| no | 30 | 65.93% | 0.6491 |
| yes | 30 | 71.32% | 0.7009 |

For the current study, we chose only $m = 2$ pairs of disputes from a sampled pair of cases. Therefore, we had 100,000 training instances for finetuning the pretrained text embedder. Each instance consisted of three parts: (a dispute item from a case of the sampled pair, a dispute item from the other case, a label of whether the pair is assumed to be similar or dis-similar).

After this preparation step, we can control the finetuning step quite precisely, using 80% and 20% of the 100,000 training instances for training and validation, respectively, and with early stopping.

### 6.2 Results: Finetuning Improved the Embedder

This subsection presents an evaluation of the finetuned `multilingual-e5-large` model's embedding quality using a simplified Similar Case Retrieval (SCR) task with 499 cases. To ensure a rigorous evaluation, we adopted a strict similarity threshold, classifying case pairs with a similarity score below 1 as "not-similar" and those with a score of 1 or above as "similar." This approach, as discussed in Section 6.1, increases the potential of confusion in the recommendations.

To assess the effectiveness of finetuning only, we employed a straightforward SCR task. Note that we did not have a classifier involved at this step. For each query case, we calculated the average Euclidean distance between its dispute items and those of all other cases (candidate cases). Candidate cases were then ranked based on this distance, with shorter distances indicating higher similarity.

While the recommender can be configured to return varying numbers of similar cases, this evaluation focused on retrieving the top 30 similar cases for each query. To facilitate this, we selected 30 query cases that had at least 30% and at most 40% of similar cases within the pool of 499, ensuring a sufficient number of relevant candidates.

Table 1 presents the evaluation results, comparing the performance of the finetuned ("FT?") and pretrained models across different numbers of retrieved cases ("topN"). Performance is measured using precision (percentage of truly similar cases within the topN recommendations) and the Normalized Discounted Cumulative Gain (NDCG).

The results shown in Table 1 demonstrate that the finetuned model consistently outperforms the pretrained model in terms of both precision and NDCG.

## 7 Evaluation Step 2: Similar Case Recommendation

As discussed in Section 5, we prepared three completely separate casesets for the SCR task. The text embedder was trained and validated with the dataset with 3380 cases. The dataset with 2630 cases was used for training and validating the recommender, and the dataset with 499 cases was used for the final evaluation in both Sections 6.2 and 7.3. These three casesets were completely disjoint, so there is no problem of data leakage.

### 7.1 A BiLSTM-Based Recommender Model

The main purpose of this evaluation is to show the usability of the results of labeling case similarity with co-citations of legal articles. We do not aim to propose a powerful method for recommending similar cases.

Fig. 8 shows the BiLSTM-based recommender that we used in the evaluation. A similar BiLSTM architecture was also used in [5]. In this model, the predictions about similarity are based on the similarity between the dispute items of the query case and a candidate case.

The dispute items of the query and the candidate cases are embedded by the same text embedder, i.e., the one we finetuned with the caseset of 3380 cases in Section 6. The embeddings are then processed by the same BiLSTM module, although the BiLSTM module is shown twice in Fig. 8. We processed the dispute items with the same BiLSTM module to ensure that the resulting vectors belong to the same vector space. The embeddings of the two cases are then fed to three layers of dense layers with different output units, as noted in the figure. The last dense layer has only one output unit, and uses a sigmoid activation function. Adding the logistic regression model step before emitting the prediction is a useful technique. This allows the logistic regression step to determine the threshold for the final layer's output. This threshold is used to classify the cases as "similar" or "not-similar."

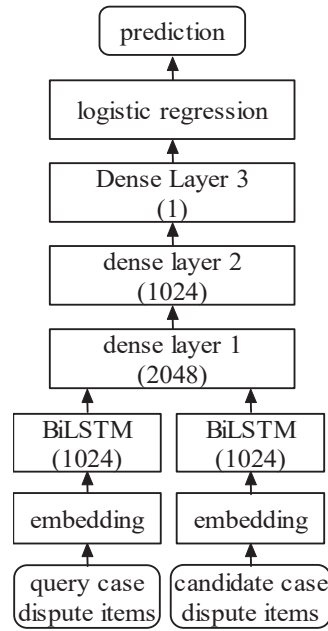

**Fig. 8.** A BiLSTM-based recommender

### 7.2 Training the Recommender

We used 2630 cases for training and validation at this step. Although we could theoretically create over 3 million case pairs,[8] we sampled only 31,124 pairs for this evaluation for data usability. This decision was based on two reasons: first, to create a balanced training and validation set (12,924 "similar" pairs and 18,200 "not-similar" pairs); and second, to avoid the "confusion" issue discussed in Section 6.1.

We employed a stratified sampling procedure to split the 31,124 pairs, allocating 80% for training and 20% for validation. A 30-fold cross-validation procedure was performed, with resampling and splitting of the data for each fold. To evaluate model performance, we used early stopping during training. The prediction results at the point of early stopping were recorded and used to generate a boxplot, allowing us to observe the effectiveness of our data in a typical machine learning context.

Fig. 9 shows the boxplot for the $F_1$ measures of the validation results that were recorded at the training phase of the BiLSTM-based recommender. Overall, the models achieved about 0.87 in the $F_1$ measures.

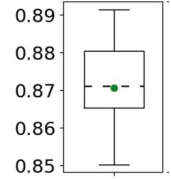

**Fig. 9.** Validating the BiLSTM-based recommender ($F_1$)

**Table 2.** Performance of the BiLSTM recommender

| topN | precision | NDCG |
|---|---|---|
| 5 | 71.18% | 0.7013 |
| 10 | 73.82% | 0.7197 |
| 15 | 75.56% | 0.7386 |
| 20 | 76.42% | 0.6900 |
| 25 | 76.78% | 0.7011 |
| 30 | 76.93% | 0.7102 |

### 7.3 Co-Citations of Legal Articles and Dispute Similarity Have Consensus

We continued to use the dataset and procedure employed in Section 6.2 to compare the recommendation quality achieved by recommenders relying on the pretrained and finetuned text embedders. For the recommender discussed in this section, we used the text embedder that was finetuned in Section 6.

Table 2 shows the performance of the BiLSTM-based recommender. The column meanings are the same as those in Table 1. We do not report the results of using a not-finetuned text embedder, so there is not a "FT?" column like Table 1.

Overall, the statistics in Table 2 indicate reasonable outcomes of an ordinary recommender. In Section 6.2, we explained that we selected 30 query cases that had at least 30% and at most 40% of similar cases within the pool of 499 cases. Hence, if our recommender was making random guesses, as governed by a binomial distribution, the probability of achieving 70% correct prediction would be no more than 0.0634%.[9] The precisions listed in Table 2 might not be very impressive, but they showed that judging similar cases based on co-citations of legal articles and based on the similarity between the disputes of the litigants have a statistically significant consensus.

## 8 Concluding Remarks and Discussions

This research demonstrated the effectiveness of leveraging co-citations of legal articles to establish similarity between labor dispute cases and enable algorithmic annotation. By finetuning text embedding models and training a basic BiLSTM model based on these automatically generated labels, we achieved promising results in recommending similar cases based on plaintiffs' accusations, defendants' rebuttals, and points of contention. Note that the co-citation of legal articles and the disputes are contextually independent information in typical judgment documents. Hence, being able to using similar disputes to recommend cases whose similarity is defined based on co-citation of legal articles is logically meaning.

---

[8] Choosing any two cases from 2630 cases have 2630×2629÷2 = 3,470,285 combinations.
[9] Given 30 tests, the probability of 70% correct recommendation is between $\binom{30}{21}(0.3)^{21}(0.6)^9$ and $\binom{30}{21}(0.4)^{21}(0.7)^9$, namely $[6.04 \times 10^{-6}, 6.34 \times 10^{-4}]$.

This approach offers a valuable solution to the challenge of limited labeled datasets in specialized legal domains, paving the way for automated annotation techniques and the development of more robust legal recommender systems, particularly in low-resource settings. Future work can focus on expanding the system to provide justifications and explanations for its recommendations, further enhancing its practical value for legal professionals.

We may have to plan how to use a huge pool of algorithmiclly labeled case pairs in our work. Sampling a reasonable amount of these algorithmically labeled case pairs for human feedback is an obvioius step. One may have noticed that the gaps between the precision values listed in Table 2 and the $F_1$ values shown in Fig. 9 were quite large. This may have been caused by the arbitrary selection of the 499 test cases. We have noticed the influence of an arbitrary selection of a test dataset from the large pool on the observed recommendation quality. Further investigation is needed to understand this phenomenum.

## Acknowledgments

This research was supported in part by the grant NSTC-113-2221-E-004-008 from the National Science and Technology Council of Taiwan. The authors were deeply indebted to the reviewers who provided invaluable comments to the previous manuscript. Although we could not respond to all of the comments completely in this paper, we would do so in a longer report that we plan to publish.